\documentclass[letterpaper, 10 pt, conference]{ieeeconf}  

\IEEEoverridecommandlockouts                              

\usepackage{xurl}
\usepackage{hyperref}
\usepackage{amsmath} 
\usepackage{amssymb}
\usepackage{tabularx}
\usepackage{xcolor}
\usepackage{svg}
\usepackage{booktabs}
\usepackage{subcaption}
\usepackage{multirow}
\usepackage[symbol]{parnotes}
\usepackage{siunitx}
\usepackage{nicematrix}
\usepackage{placeins}
\usepackage{array}
\usepackage{booktabs}
\usepackage{makecell}
\usepackage{tabularx}
\usepackage{comment}

\definecolor{mylog}{HTML}{636363}
\definecolor{mycarl}{HTML}{1F77B4}
\definecolor{mydiff}{HTML}{ff7f0e}
\definecolor{mypluto}{HTML}{d62728}
\definecolor{myplantf}{HTML}{9467bd}
\definecolor{mypdmc}{HTML}{2ca02c}

\newif\ifanonymous
\anonymousfalse 

\title{\LARGE \bf
Shift \& Drift: A Zero-Shot Benchmark for Generalizable and Robust\\Autonomous Driving Motion Planning
}

\ifanonymous
    \makeatletter
    \renewcommand\thanks[1]{}
    \makeatother

  \author{Anonymous Authors}
\else
    \author{Alessandro Canevaro$^{1,2,\ast,\dag}$, Hang Yu$^{1,3, \ast}$, Julian Schmidt$^{1}$, Peizheng Li$^{1,2}$,\\Silvan Lindner$^{1}$, Wilhelm Stork$^{3}$, Georg Martius$^{2}$, and Julian Jordan$^{1}$ %
    \thanks{This work is a result of the joint research project STADT:up (19A22006O). The project is supported by the German Federal Ministry for Economic Affairs and Energy (BMWE), based on a decision of the German Bundestag. The authors are solely responsible for the content of this publication.
    Code and data are available at:\newline    
    \url{https://github.com/alessandro-canevaro/Shift-Drift}.}    
    \thanks{$^{1}$Mercedes-Benz AG, Research \& Development, Sindelfingen, Germany}%
    \thanks{$^{2}$University of T{\"u}bingen, T{\"u}bingen, Germany}%
    \thanks{$^{3}$Karlsruhe Institute of Technology, ITIV, Karlsruhe, Germany}%
    \thanks{$^{\dag}${\tt\small alessandro.canevaro@mercedes-benz.com}}%
    \thanks{$^\ast$Equal contribution.}
    }
\fi

\begin{document}

\maketitle
\thispagestyle{empty}
\pagestyle{empty}

\begin{abstract}

While closed-loop motion planners trained on large-scale, object-level datasets, e.g., nuPlan, demonstrate strong in-distribution (ID) performance, their generalization to novel urban topologies and recovery mechanisms following execution perturbations remain under-explored.
To address this, we present \textit{Shift \& Drift}, a novel dual-track benchmark designed to rigorously stress-test motion planners across two critical axes of distribution shift:
(1) The \textit{Semantic Shift Track} leverages a novel conversion pipeline that transforms the aerial, DeepScenario Open 3D dataset into the nuPlan simulation framework.
This enables zero-shot evaluation of planners trained on North American and Singaporean data against 1,182 scenarios spanning four German cities and the US city of San Francisco featuring dense pedestrian-cyclist interactions.
(2) The \textit{State-Distribution Drift Track} injects stochastic perturbations into the ego vehicle’s dynamics to quantify robustness against compounding execution errors.
Based on this, we systematically evaluate the failure modes of diverse planning paradigms under semantic and state-distribution shifts.
While imitation learning methods achieve high scores in ID benchmarks, they exhibit significant failures under semantic shift, particularly in pedestrian-dense environments, and suffer from persistent drift when subjected to temporally correlated actuation noise. 
In contrast, the evaluated reinforcement-learning-based planner demonstrates more graceful degradation, maintaining higher safety and progress metrics across both tracks.
Our findings reveal an empirical trade-off between imitation fidelity and closed-loop resilience, providing the community with a rigorous benchmark to evaluate progress toward reliable deployment.

\end{abstract}

\section{Introduction}
The advancement of learning-based Autonomous Driving (AD) planners has been primarily driven by large-scale datasets and standardized simulators~\cite{survey_dl_ad}. 
However, a significant gap remains between performance in familiar training environments and reliability in the real world. 
Current State-of-the-Art (SOTA), Imitation Learning (IL), Reinforcement Learning (RL), and rule-based planners, are predominantly evaluated on the same geographic distributions found in their training sets~\cite{BADUE2021113816, 10.1109/TITS.2021.3134702}. 
This reliance on i.i.d. (independent and identically distributed) data assumptions often masks a critical lack of generalization. 
As noted by Codevilla et al. \cite{CodevillaBehaviorCloning}, models that excel in known urban layouts often suffer from geographic overfitting, where the policy implicitly memorizes map-specific features or localized traffic patterns rather than learning universal driving priors.
This challenge is further compounded by causal confusion \cite{de2019causal}, where planners may latch onto spurious correlations rather than transferable driving priors.
Similarly, rule-based planners can face scalability challenges when cost-function tuning must account for diverse long-tail scenarios.

\begin{figure}[t]
    \centering
    \includegraphics[width=\linewidth]{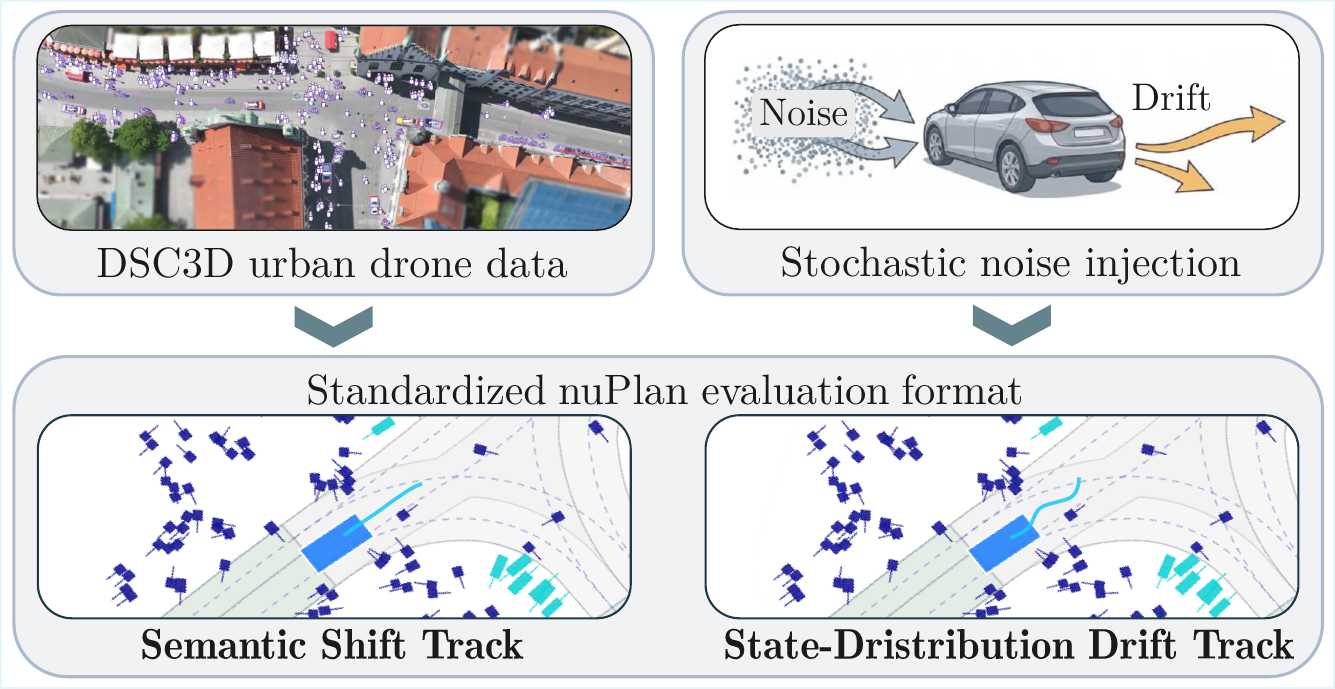}
    \caption{Overview of our dual-track benchmark: \textit{Shift \& Drift}. Track 1: Semantic Shift Track converts aerial DSC3D data~\cite{deepscenario3d2025} into nuPlan~\cite{nuplan} standard for zero-shot evaluation. Track 2: State-Distribution Drift Track injects different types of noise into actuation.}
    \vspace{-1.3em}
    \label{fig:teaser}
\end{figure}

Specifically, a manifestation of these challenges is the significant degradation in planning capability when models encounter semantic shift, such as novel urban layouts or intricate multi-agent interactions featuring high-density pedestrians and cyclists.
While benchmarks like nuPlan \cite{nuplan} and Waymo open dataset \cite{xu2025waymoopendataset} have scaled the volume of data, they still fall short of covering all potential edge cases~\cite{8954462, 10509812}, particularly complex, high-density
interaction scenarios.

Furthermore, even when a planner exhibits high fidelity in nominal conditions, it often lacks robustness against state-distribution shift. 
This compounding error problem, fundamentally identified in~\cite{NIPS1988_812b4ba2}, was later formalized in the work~\cite{pmlr-v15-ross11a}.
It is observed that a model trained strictly on expert trajectories lacks the recovery behaviors necessary to handle the drift caused by small, cumulative execution error.
While closed-loop evaluation has become the standard for AD planners, current benchmarks like CARLA~\cite{Dosovitskiy17} or nuPlan often utilize simplified vehicle dynamics models. 
In real-world execution, errors in the dynamic model, sensor latency, and stochastic actuation errors are inevitable. 
Without mechanisms to force a planner to recover from these off-policy states, closed-loop success scores remain an optimistic upper bound rather than a measure of true deployment readiness.

In this paper, we introduce \textit{Shift \& Drift}, a new benchmark designed to stress-test the generalization and robustness of motion planners. 
Our benchmark, as shown in Fig.~\ref{fig:teaser}, focuses on two primary axes of distribution shift relevant for safety-critical deployment:

\noindent\textbf{Semantic Shift through Cross-Dataset Generalization.}
We present DeepPlan, a suite of nuPlan-compatible scenarios created by mapping the DeepScenario Open 3D Dataset (DSC3D) \cite{deepscenario3d2025} into the nuPlan framework.
This allows us to leverage the high-precision, occlusion-free, drone-recorded aerial trajectories of DSC3D, captured across five diverse international locations, to conduct a rigorous zero-shot evaluation of models trained on the original nuPlan dataset.
Beyond quantifying policy generalization versus map memorization, this benchmark evaluates planner performance in more challenging, high-density environments characterized by extensive pedestrian-cyclist interactions.

\noindent\textbf{State-Distribution Drift through Dynamic Noise Injection.} 
We formalize the state distributional drift by injecting stochastic perturbations into the ego-vehicle’s transition dynamics during closed-loop simulation. 
To model different noise characteristics, we employ Additive White Gaussian Noise (AWGN) to represent high-frequency uncorrelated jitter and the Ornstein-Uhlenbeck (OU) process~\cite{PhysRev.36.823} to capture low-frequency systematic errors.
This forces the planner to recover from off-policy states, providing a quantitative measure of robustness and recovery capability.

We evaluate a comprehensive suite of SOTA planners, including PDM-Closed~\cite{Dauner2023CORL}, PlanTF~\cite{cheng2023plantf}, PLUTO~\cite{cheng2024pluto}, Diffusion Planner~\cite{zheng2025diffusionbased} and CaRL~\cite{Jaeger2025CoRL}. 
Our results provide empirical evidence for the inherent trade-offs between different learning paradigms: while IL-based models often excel in nominal conditions, they exhibit significant fragility in both evaluation tracks, with overall score performance dropping up to 76\% in novel urban environments. 
Rule-based models also drop in performance significantly, typically with a decrease in progress, but without compromising safety, achieving the lowest number of collisions.
Conversely, our evaluation shows that CaRL, an RL-based method, exhibits superior resilience across all tracks, in particular exhibiting at most an 8\% performance decay even under high-intensity state-distribution drift.

\section{Related Work}
\label{sec:related_work}

\subsection{Large-Scale Autonomous Driving Datasets}
Large-scale datasets serve as the cornerstone for modern learning-based motion planners. 
While early benchmarks primarily supported perception tasks~\cite{Geiger2012CVPR}, recent datasets such as Argoverse 2~\cite{Argoverse2}, Waymo Open~\cite{xu2025waymoopendataset} and nuPlan~\cite{nuplan} provide the large-scale, multi-agent trajectories required for planning. 
Recent efforts like ScenarioNet~\cite{li2023scenarionet} have further expanded this landscape by providing a unified platform to aggregate these heterogeneous datasets into a common simulation format.
However, these datasets are derived from ego-centric sensor logs, which are inherently subject to occlusion~\cite{9294301, doi:10.1177/03611981231185768}.

While occlusion is often treated as a perception nuisance, it introduces a fundamental challenge in closed-loop simulation. 
When a planner deviates from the recorded expert trajectory, an inevitability in closed-loop evaluation, it may enter spatial regions that were blind spots for the original ego-vehicle. 
In these off-policy states, the simulator may fail to account for actors that were physically present but occluded during the initial data capture. 
Consequently, the agent may navigate through phantom empty space, leading to an overestimation of safety and performance. 

Aerial datasets such as highD~\cite{highDdataset}, INTERACTION~\cite{interaction2019}, DSC3D~\cite{deepscenario3d2025}, and DeepUrban~\cite{selzer2024deepurban} mitigate this problem by offering occlusion-free trajectories.
However, they are predominantly designed for trajectory forecasting tasks and they lack the standardized simulation interfaces and map formats required for closed-loop planning evaluation.

Our work transforms the DSC3D dataset into the nuPlan framework, ensuring that the multi-agent interaction context remains complete regardless of the ego-vehicle’s displacement, providing a more rigorous and realistic constraint on the planner’s behavior.

\subsection{Motion Planning Paradigms}
Current SOTA motion planners can be broadly categorized into three paradigms, each with distinct trade-offs regarding generalization and robustness. 

IL paradigms, including the transformer-based PlanTF~\cite{cheng2023plantf} and the generative Diffusion Planner~\cite{zheng2025diffusionbased}, leverage human data but are limited by their open-loop nature and causal confusion~\cite{de2019causal}, often resulting in a lack of robust recovery behaviors.
Hybrid models like PLUTO~\cite{cheng2024pluto} attempt to bridge this gap using a rule-based refinement layer for traffic law adherence. 
While interactive techniques like DAgger~\cite{pmlr-v15-ross11a} address distributional drift via expert querying, they scale poorly to real-world datasets where experts are unavailable post-recording.

RL approaches, such as CaRL~\cite{Jaeger2025CoRL}, a SOTA nuPlan agent trained via PPO~\cite{schulman2017ppo} to maximize progress, safety, and comfort, mitigate these shifts through closed-loop exploration. 
However, RL remains computationally expensive and requires extensive reward engineering due to poor sample efficiency.

Finally, rule-based planners like PDM-Closed~\cite{Dauner2023CORL} serve as deterministic baselines, utilizing IDM~\cite{Treiber_2000} and MOBIL~\cite{mobil} for longitudinal and lateral control. 
While providing strong safety guarantees and interpretability, they require exhaustive tuning and often fail to generalize to complex long-tail multi-agent interactions.

\subsection{Closed-Loop Benchmarking and Generalization}
Pioneered by simulators like CARLA~\cite{Dosovitskiy17} and benchmarks including Bench2Drive~\cite{jia2024bench}, nuPlan~\cite{nuplan} and NAVSIM~\cite{Cao2025CORL}, the transition from open-loop displacement metrics to closed-loop simulation has become the standard for evaluating sequential decision-making. 
Recent advancements have focused on improving the realism of background agents. 
For example, nuPlan-R~\cite{peng2025nuplanrclosedloopplanningbenchmark} and the integration of SMART agents \cite{hagedorn2025plannersmeetrealitylearned} replace standard rule-based models with learned, reactive agents, complementing our work by improving the realism of simulated traffic agents.
However, most existing benchmarks rely on i.i.d. evaluation, where models are tested on held-out data from the same geographic and semantic distribution as the training set. 
This often masks a critical lack of generalization, as models may implicitly memorize map-specific features or localized traffic patterns.

A recent effort to address these deficiencies is interPlan \cite{Hallgarten2024interPlan}, which challenges the optimistic upper bound of the nuPlan benchmark. 
This refers to the observation that many planners achieve high scores in nuPlan by excelling at standard cruising tasks in simple scenarios, failing to reveal their fragility in complex interactions. 

While interPlan effectively probes behavioral reliability in rare scenarios, its 80 handcrafted cases do not provide the statistical scale required for comprehensive validation.
Our work advances this frontier by enabling large-scale cross-dataset evaluation. 
By converting 1,100+ diverse real-world recordings from DSC3D, we move beyond manual scenario design to provide a data-driven measure of semantic generalization across international geographic domains (see Table~\ref{tab:benchmark_comparison} for a comparative analysis).

\subsection{Robustness and Sim-to-Real Gaps}
The disparity between simulation and reality, often termed the sim-to-real gap, remains a critical hurdle.
Prior work explored this through adversarial attacks \cite{xu2022safebench, 9578745} or by varying visual domains in perception-based pipelines.
However, for object-based planning, a subtle but critical shift arises from imperfect low-level control.
While robust control theory is well-established, its integration into AD planning benchmarks is under-explored.
Unlike adversarial approaches seeking worst-case perturbations, our State-Distribution Drift Track utilizes Gaussian and temporally-correlated noise models to simulate execution errors, providing a practical proxy for a planner's stability on imperfect hardware.

\begin{table}[tb]
\centering
\caption{Comparison of motion planning benchmarks.}
\label{tab:benchmark_comparison}
\footnotesize
\resizebox{\columnwidth}{!}{
\begin{tabular}{lccc}
\toprule
\textbf{Attribute} & \textbf{nuPlan} & \textbf{InterPlan} & \textbf{DeepPlan (Ours)} \\
\midrule
Data Source & Vehicle & Vehicle & Aerial \\
Occlusion-Free Tracking & $\times$ & $\times$ & \checkmark \\
Geographic Coverage & US + SG & US + SG & US + DE \\
Semantic Shift & Limited & Scenario & \checkmark \\
Interaction Density & Moderate & High & Very High \\
Train--Test Overlap Risk & High & Moderate & Low \\
Scenario Count & 1118 & 80 & 1182\\
\bottomrule
\end{tabular}
}
\vspace{-1.3em}
\end{table}

\section{The Benchmark}
Our benchmark is structured into two independent evaluation tracks designed to probe the critical limitations of current AD planners along two distinct axes: semantic generalization and robustness to state-distribution shifts. 

\subsection{The Semantic Shift Track}
The foundation of this track is DeepPlan, a novel benchmark generated via a pipeline that converts the high-precision, real-world aerial DSC3D~\cite{deepscenario3d2025} data into a format compatible with the standardized nuPlan~\cite{nuplan} simulator.
Thus, models trained exclusively on the North American and Singaporean locations within the nuPlan dataset can be evaluated on zero-shot basis in newly converted urban scenes such as Munich, Stuttgart, and others.
This track is designed to answer a critical question: have planners learned generalizable driving policies, or have they merely memorized the topology and traffic patterns of their training distribution?

The DSC3D scenarios present a significant semantic shift, characterized by:
\begin{itemize}
    \item \textbf{Novel Urban Topologies:} Planners must navigate road layouts and intersection types not present in the original nuPlan dataset like pedestrian areas and parking lots.
    \item \textbf{Dense Social Interactions:} The high density of pedestrians and cyclists in the selected scenarios significantly exceeds that of standard nuPlan logs, as illustrated in Fig.~\ref{fig:data_stats}, thereby testing the capacity of planners for complex multi-agent reasoning and safe navigation among vulnerable road users.
    \item \textbf{Regional Driving Styles:} The evaluation exposes planners to distinct and localized driving behaviors, and norms inherent to European locations.
\end{itemize}

Our conversion process transforms the DSC3D raw data into nuPlan-compatible log files and vector maps through a multi-stage pipeline:

\subsubsection{\textbf{Vector Map Conversion}} We process the accompanying OpenDRIVE~\cite{asam_opendrive} map files for each DSC3D location to extract semantic and geometric primitives (lane boundaries, centerlines, roadblocks, speed limits).
This process includes: (i) automatic intersection detection via spatial overlap analysis of lane polygons, (ii) coordinate system alignment from GPS/OpenDRIVE to UTM projections with ego pose registration, and (iii) R-tree spatial indexing for efficient map queries during simulation.
Outputs are layered in GeoPackage format required by nuPlan, ensuring planners use the same high-level semantic representations consistent with their training.

\subsubsection{\textbf{Episode Windowing and Ego Vehicle Selection}} 
Raw 12.5 Hz DSC3D recordings are segmented into 18-second sliding windows (225 frames) with a 6-second overlap to maximize data coverage while maintaining temporal diversity.
Since DSC3D is a drone-recorded, ego-less dataset, a candidate ego must be identified programmatically. To qualify, every tracked agent is evaluated against the following criteria: (i) it must be classified as a vehicle (excluding pedestrians, cyclists, and trucks); (ii) its peak speed over the episode must exceed 1.0 m/s; and (iii) it must not experience any bounding-box overlap with surrounding agents. 
Each qualifying vehicle in a window yields an independent nuPlan database file, so a single window can produce multiple episodes. 
Each episode is then upsampled to nuPlan's 20 Hz via linear interpolation, producing 360-frame episodes.
Within each episode, frame 60 is defined as the current timestep ($t=0$) delineating 3 seconds of history and 15 seconds of future context.

\subsubsection{\textbf{Agent States and Log Generation}} For each episode, 
we compute comprehensive kinematic states (position, velocity, acceleration, heading) for all tracked agents. These states, coupled with ego odometry, are serialized into the nuPlan SQLite schema for closed-loop replay.

\subsubsection{\textbf{Post-Processing}} The raw conversion output is subject to two stages of post-processing:
(i) an automated filter evaluates each generated scenario under the nuPlan closed-loop simulator using the log playback planner, and discards database file where the ego exhibits at-fault collisions, drivable-area violations, or insufficient forward progress, indicating a malformed or degenerate scenario. 
(ii) a manual inspection tool renders each remaining scenario with its semantic map layers and recorded ego trajectory, allowing a human reviewer to visually inspect individual scenarios and to correct route roadblock assignments where necessary.

\begin{figure}[t]
    \centering
    \includegraphics[width=\linewidth]{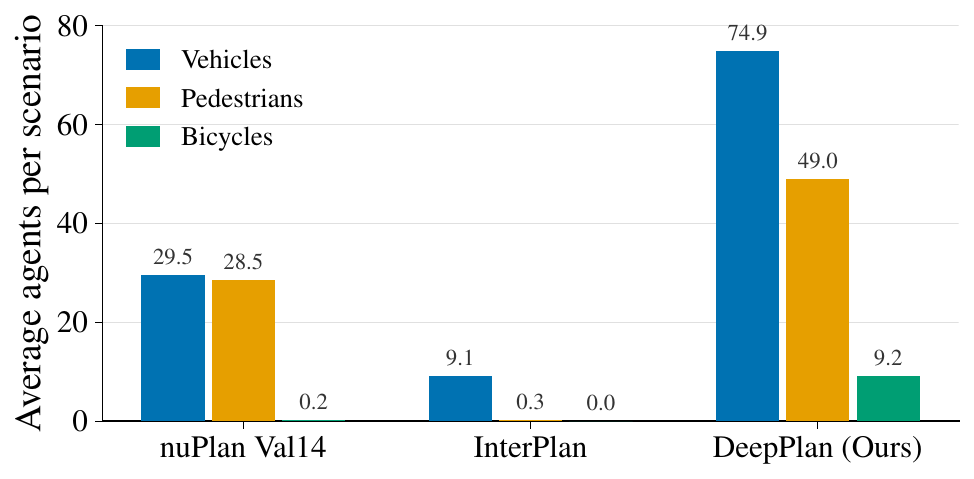}
    \caption{Average number of vehicles, pedestrian and bicycles per scenario on nuPlan val14, InterPlan and DeepPlan.}
    \vspace{-1.3em}
    \label{fig:data_stats}
\end{figure}

\noindent\textbf{Dataset Characteristics:}
The finalized evaluation set comprises 1182 scenarios distributed as follows: 205 for Munich, 206 for Stuttgart, 490 for Sindelfingen, 101 for Berlin and 180 for San Francisco.
Fig.~\ref{fig:data_stats} reports the average number of vehicles, pedestrians and bicycles in the nuPlan Val14, InterPlan and DeepPlan datasets.

\subsection{The State-Distribution Drift Track}
\label{sec:noise_levels}

To quantify the robustness of the evaluated planners against compounding errors, a phenomenon where small execution inaccuracies push the agent into out-of-distribution states, we introduce the State-Distribution Drift Track.
Rather than faithfully replicating specific hardware imperfections, we employ controlled stochastic models as standardized stress-test proxies.
We perturb the ego-vehicle's command vector $\mathbf{u}_\text{cmd} = [a_\text{cmd},\, \dot{\delta}_\text{cmd}]^\top$, where $a_\text{cmd}$ is the longitudinal acceleration command and $\dot{\delta}_\text{cmd}$ is the steering rate command, using two distinct stochastic mechanisms.

\subsubsection{\textbf{Stochastic Actuation Perturbations}}
We model high-frequency mechanical jitter and control inaccuracies by injecting Additive White Gaussian Noise (AWGN) directly into the command outputs:
\begin{equation}
    \begin{bmatrix} a_\text{applied} \\ \dot{\delta}_\text{applied} \end{bmatrix}
    =
    \begin{bmatrix} a_\text{cmd} \\ \dot{\delta}_\text{cmd} \end{bmatrix}
    + \boldsymbol{\varepsilon},
    \quad
    \boldsymbol{\varepsilon} \sim \mathcal{N}\!\left(\mathbf{0},\,
    \mathrm{diag}\!\left(\sigma_a^2,\, \sigma_{\dot{\delta}}^2\right)\right).
\end{equation}

Because $\boldsymbol{\varepsilon}$ is drawn independently at each timestep, errors are temporally uncorrelated.
This tests the planner's ability to maintain a smooth trajectory despite instantaneous, high-frequency jitter.
We evaluate three intensity levels:

\begin{itemize}
    \item Low:  $\sigma_{\dot{\delta}} = 0.1\,\text{rad/s}$,\; $\sigma_{a} = 0.5\,\text{m/s}^2$.
    \item Mid:  $\sigma_{\dot{\delta}} = 0.2\,\text{rad/s}$,\; $\sigma_{a} = 1.0\,\text{m/s}^2$.
    \item High: $\sigma_{\dot{\delta}} = 0.3\,\text{rad/s}$,\; $\sigma_{a} = 1.5\,\text{m/s}^2$.
\end{itemize}

\subsubsection{\textbf{Correlated Drift}}
To simulate realistic, low-frequency systematic errors, such as wheel misalignment or environmental forces like crosswinds, we perturb the planner's commanded acceleration $a_\text{cmd}$ and steering rate $\dot{\delta}_\text{cmd}$ using two independent Ornstein-Uhlenbeck (OU) processes \cite{PhysRev.36.823}.
Given the noise vector $\mathbf{x}_k = [x_k^{a},\, x_k^{\dot{\delta}}]^\top$, the applied commands are:
\begin{equation}
    \begin{bmatrix} a_\text{applied} \\ \dot{\delta}_\text{applied} \end{bmatrix}
    =
    \begin{bmatrix} a_\text{cmd} \\ \dot{\delta}_\text{cmd} \end{bmatrix}
    + \mathbf{x}_k.
\end{equation}

Each component $x_k^{(\cdot)}$ evolves according to the continuous-time OU SDE
\begin{equation}
    dx_t = \theta\,(\mu - x_t)\,dt + \sigma\,dW_t,
\end{equation}

which is integrated numerically at every simulation step $\Delta t$ via the Euler-Maruyama scheme:
\begin{equation}
    x_{k+1} = x_k + \theta\,(\mu - x_k)\,\Delta t
              + \sigma\,\sqrt{\Delta t}\;\varepsilon_k, \, \varepsilon_k \sim \mathcal{N}(0, 1).
\end{equation}

The two processes share the same mean-reversion rate $\theta = 2.0\,\text{s}^{-1}$ and long-term mean $\mu = 0$, but differ in volatility, using the same three $\sigma$ intensity levels as the AWGN track for direct comparison.
The mean-reversion term $\theta(\mu - x_k)\Delta t$ prevents unbounded random-walk behavior, while the stochastic term $\sigma\sqrt{\Delta t}\,\varepsilon_k$ injects new randomness scaled to preserve the correct per-unit-time variance.
The resulting noise exhibits an autocorrelation that decays as $e^{-\theta\tau}$, yielding a characteristic correlation time of $1/\theta = 0.5\,\text{s}$. 
This timescale is sufficiently long to produce multi-frame drifts that represent a genuine disturbance (see Fig.~\ref{fig:qual_noise}.4), yet bounded in expectation. 
Consequently, the planner’s receding-horizon feedback must actively compensate for accumulated error rather than merely reacting to instantaneous perturbations.

\subsection{Evaluation Metrics}
We adopt the standard closed-loop evaluation protocol of nuPlan, which assesses performance across safety, rule compliance, progress, and passenger comfort. 
The metric suite includes collision avoidance, driving direction compliance, drivable area adherence, time-to-collision (TTC), speed limit compliance, and comfort metrics such as acceleration and jerk. 
Following the original framework, safety-critical violations act as hard constraints where exceeding predefined thresholds results in a scenario score of zero, while the remaining metrics are aggregated into a weighted average to compute the final score.

For the DeepPlan evaluation, we adjust a subset of these thresholds (Table~\ref{tab:metric_changes}) to account for the unique characteristics of dense, pedestrian-heavy European environments. 
Unlike the predominantly vehicle-centric, wide-lane traffic in nuPlan's North American logs, DeepPlan scenarios feature significantly higher interaction density and frequent negotiation of shared spaces. 
In these settings, safe, human-like navigation can often require tighter TTC margins or temporary deviations from nominal lane boundaries during yielding maneuvers.

\begin{table}[tb]
\centering
\caption{Modified nuPlan evaluation thresholds for DeepPlan scenarios.}
\label{tab:metric_changes}
\footnotesize
\begin{tabular}{lcc}
\toprule
\textbf{Metric} & \textbf{Default} & \textbf{Ours} \\
\midrule
Driving Direction Compliance (m) & 2.0 & 4.0 \\
Driving Direction Violation (m) & 6.0 & 10.0 \\
Drivable Area Violation (m) & 0.3 & 3.0 \\
Max Longitudinal Jerk ($m/s^3$) & 4.13 & 6.0 \\
Min Longitudinal Accel. ($m/s^2$) & -4.05 & -6.0 \\
Max Longitudinal Accel. ($m/s^2$) & 2.40 & 4.0 \\
Minimum TTC (s) & 1.0 & 0.5 \\
\bottomrule
\end{tabular}
\vspace{-1.3em}
\end{table}

To validate these adjustments, we conducted a sanity check by evaluating the original log playback (the human driver) using the default nuPlan thresholds on the DeepPlan set. 
The log's score dropped from approximately 90 to 65, primarily due to spurious safety violations in scenarios that were, in reality, safe human maneuvers. 
Applying the default thresholds would therefore systematically penalize reasonable behavior rather than actual planning failures. 
Compared to existing benchmarks like interPlan that modify the underlying metric logic, our approach maintains the original nuPlan scoring equations, requiring only adjusting threshold values to ensure the benchmark remains physically meaningful across geographic domains. 
\section{Experiments}
\label{sec:experiments}

\begin{table*}[bt]
\centering
\caption{\textbf{Zero-Shot Generalization Performance.} Comparison of planners on the standard nuPlan validation set (Val14) versus our proposed semantic shift benchmark.}
\label{tab:semantic_shift}

\resizebox{\textwidth}{!}{
\begin{tabular}{l | c | cccccc | cccccccc}
\toprule

&
\multicolumn{1}{c|}{\textbf{nuPlan}} &
\multicolumn{6}{c|}{\textbf{Cities (CLS)}} &
\multicolumn{8}{c}{\textbf{Metrics}} \\

\cmidrule(lr){2-2}
\cmidrule(lr){3-8}
\cmidrule(lr){9-16}

\textbf{Method}
& \textbf{Val14}
& \textbf{All} & \textbf{SIFI} & \textbf{MUC} & \textbf{STR} & \textbf{BER} & \textbf{SFO}
& \textbf{DAC} & \textbf{DDC} & \textbf{EIC} & \textbf{EMP} & \textbf{PER} & \textbf{NCR} & \textbf{SLC} & \textbf{TTC} \\

\midrule
\multicolumn{16}{c}{\textbf{Non-Reactive}} \\
\midrule

Log  & 96.39 & 90.71 & 88.35 & 90.00 & 96.95 & 85.03 & 93.97 & 100.00 & 94.83 & 90.77 & 99.83 & 99.13 & 100.00 & 99.97 & 90.77 \\
PDM-Closed & 92.84 & 48.07 & 27.22 & 57.02 & 59.59 & 54.58 & 77.77 & 98.81 & 99.11 & 72.17 & 66.67 & 50.21 & 89.25 & 99.99 & 77.75 \\
PlanTF    & 84.27 & 34.00 & 28.84 & 29.24 & 40.58 & 44.14 & 40.28  & 98.81 & 98.77 &  78.93 & 52.36 & 36.22 & 86.04 & 99.98 & 78.34 \\
PLUTO     & 92.88 & 36.74 & 11.94 & 45.57 & 55.44
& 61.98
& 59.53 & 99.82 & 99.51 & 69.89 & 52.69 & 39.14 & 88.98 & 99.99 & 80.91 \\
Diffusion planner & 89.87 & 37.49 & 49.16 & 21.50 & 31.57 & 26.57 & 36.80 & 97.63 & 97.29 & 61.92 & 63.28 & 48.65 & 78.93 & 99.97 & 74.11 \\
CaRL      & 93.87 & 72.57 & 73.51 & 57.59 & 72.73 & 73.33 & 86.47 & 99.74 & 97.03 & 77.66 & 91.37 & 83.29 & 89.25 & 99.98 & 78.08 \\

\midrule
\multicolumn{16}{c}{\textbf{Reactive}} \\
\midrule

Log  & 82.02 & 71.63 & 66.61 & 78.36 & 68.50 & 58.22 & 88.75 & 100.00 & 94.83 & 90.77 & 99.83 & 99.13 & 78.25 & 99.97 & 73.35 \\
PDM-Closed & 92.12 & 49.95 & 25.27 & 60.72 & 65.45
& 55.40 & 84.05 & 97.80 & 98.94 & 72.08 & 67.85 & 53.03 & 90.10 & 99.99 & 80.29 \\
PlanTF  & 76.95 & 49.44 & 42.57 & 38.22 & 55.98 & 57.47
& 68.90  & 96.70 & 95.35 & 83.08 & 75.21 & 58.71 & 83.76 & 99.99 & 76.31 \\
PLUTO     & 92.06 & 63.03 & 50.27 & 62.69 & 66.42 & 76.83 & 88.28 & 99.91 & 99.09 & 83.22 & 82.58 & 60.57 & 89.33 & 99.99 & 80.67 \\
Diffusion planner & 82.80 & 49.39 & 48.28 & 34.30 & 62.59 & 41.36 & 59.00 & 89.84 & 94.54 & 69.03 & 90.10 & 74.33 & 71.99 & 99.99 & 66.92 \\
CaRL      & 93.12 & 70.27 & 61.19 & 68.90 & 76.04 & 74.29 & 87.67 & 99.83 & 96.61 & 74.78 & 93.23 & 84.12 & 84.43 & 99.99 & 77.74 \\

\bottomrule
\end{tabular}
}
\vspace{-1.0em}
\end{table*}

\subsection{Experimental Setup}
\subsubsection{Baselines}
To ensure a comprehensive evaluation across the current planning landscape, we select five SOTA planners representing three dominant paradigms:

\noindent\textbf{Rule-Based.}
We include the nuPlan challenge winner PDM~\cite{Dauner2023CORL}.

\noindent\textbf{Imitation Learning.}
We compare against PlanTF~\cite{cheng2023plantf}, PLUTO~\cite{cheng2024pluto}, and Diffusion Planner~\cite{zheng2025diffusionbased}.

\noindent\textbf{Reinforcement Learning.}
We evaluate CaRL~\cite{Jaeger2025CoRL}, a recent SOTA closed-loop learning approach.

\subsubsection{Metrics}
Adhering to the standard nuPlan evaluation protocol, we assess planner performance across several metrics. 
The final performance is distilled into a composite \textbf{Closed-Loop Score (CLS)}, representing the weighted mean of the following:

\noindent\textbf{Drivable Area Compliance (DAC):} Quantifies adherence to road boundaries and penalizes off-road excursions.

\noindent\textbf{Driving Direction Compliance (DDC):} Penalizes driving against the flow of traffic.

\noindent\textbf{Ego is Comfortable (EIC):} Measures kinematic smoothness by penalizing excessive jerk and centripetal acceleration.

\noindent\textbf{Ego is Making Progress (EMP):} Binary metric validating minimum 20\% progress relative to the expert.

\noindent\textbf{Ego Progress Along Expert Route (PER):} Measures the ratio of ego-to-expert progress along the designated route.

\noindent\textbf{Non-Collision Rate (NCR):} Safety metric calculating the percentage of scenarios completed without at-fault collision.

\noindent\textbf{Speed Limit Compliance (SLC):} Penalizes any exceedance of the posted speed limit.

\noindent\textbf{Time To Collision (TTC):} Measures the temporal buffer to lead agents to penalize near-misses and tailgating.

In the following sections we report results on the two nuPlan simulation modes: Non-Reactive (NR) where agents are replayed from the log, and Reactive (R) where neighbor vehicles are controlled by the IDM model.

\begin{figure*}[hbt]
    \centering
    \includegraphics[width=\textwidth]{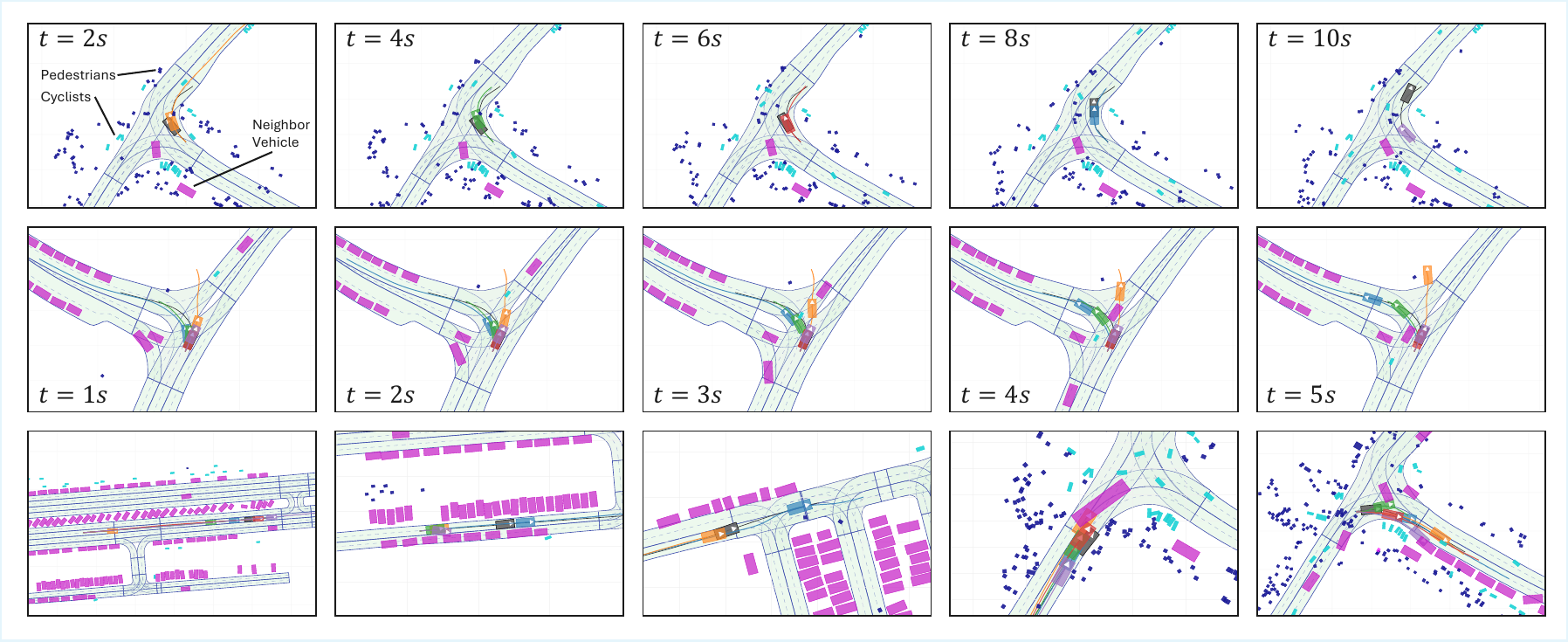}
    \caption{\textbf{Qualitative comparison of planner trajectories across DeepPlan scenarios.} 
    We show the ground truth (\textbf{\textcolor{mylog}{Log}}) alongside \textbf{\textcolor{myplantf}{PlanTF}}, \textbf{\textcolor{mypdmc}{PDM-Closed}}, \textbf{\textcolor{mypluto}{PLUTO}}, \textbf{\textcolor{mydiff}{Diffusion Planner}}, and \textbf{\textcolor{mycarl}{CaRL}}. 
    \textbf{Top row:} During a right turn with high pedestrian density, Diffusion Planner and PDM-Closed proceed aggressively, resulting in collisions, while PLUTO and CaRL successfully yield. PlanTF fails to commit to forward progress.
    \textbf{Middle row:} In an unprotected left turn, several planners fail to account for oncoming traffic. 
    \textbf{Bottom row (left to right):} Diffusion Planner exhibits freezing behavior on an empty road; most planners fail to overtake parked vehicles using the opposite lane, whereas CaRL successfully navigates the obstacle; CaRL invades opposite lane to avoid pedestrian; all planners struggle to predict bus motion, leading to suboptimal stopping; finally, extreme urban density causes universal failure across all models, highlighting the complexity of the DeepPlan environment.
    }
    \vspace{-1.0em}
    \label{fig:track1_qualitative}
\end{figure*}

\begin{table*}[t]
\centering
\caption{\textbf{Noise Robustness Performance.} Comparison of planners on nuPlan (Val14) and DeepPlan validation sets with different type and level of noise on dynamic model. Noise levels and OU parameters are defined in Section~\ref{sec:noise_levels}. Results are reported as $\text{mean}_{\text{std}}$ over three random seeds.}
\label{tab:noise_comparison}
\resizebox{\textwidth}{!}{
\begin{tabular}{l | c | ccc | ccc | c | ccc | ccc}
\toprule
& \multicolumn{7}{c|}{\textbf{nuPlan}} & \multicolumn{7}{c}{\textbf{DeepPlan}} \\
\cmidrule(lr){2-8} \cmidrule(lr){9-15}
& & \multicolumn{3}{c|}{\textbf{AWGN}} & \multicolumn{3}{c|}{\textbf{OU}} & & \multicolumn{3}{c|}{\textbf{AWGN}} & \multicolumn{3}{c}{\textbf{OU}} \\
\cmidrule(lr){3-5} \cmidrule(lr){6-8} \cmidrule(lr){10-12} \cmidrule(lr){13-15}
\textbf{Method} & \textbf{Val14} & \textbf{Low} & \textbf{Mid} & \textbf{High} & \textbf{Low} & \textbf{Mid} & \textbf{High} & \textbf{Base} & \textbf{Low} & \textbf{Mid} & \textbf{High} & \textbf{Low} & \textbf{Mid} & \textbf{High} \\
\midrule
\multicolumn{15}{c}{\textbf{Non-Reactive}} \\
\midrule
PDM-Closed & 92.84 & 90.92\textcolor{gray}{\tiny{0.60}} & 87.21\textcolor{gray}{\tiny{2.73}} & 76.46\textcolor{gray}{\tiny{3.29}} & 88.49\textcolor{gray}{\tiny{2.87}} & 80.05\textcolor{gray}{\tiny{3.36}} & 70.45\textcolor{gray}{\tiny{3.54}} & 48.07 & 46.91\textcolor{gray}{\tiny{0.40}} & 44.90\textcolor{gray}{\tiny{2.75}} & 41.37\textcolor{gray}{\tiny{1.95}} & 45.74\textcolor{gray}{\tiny{0.31}} & 43.46\textcolor{gray}{\tiny{0.37}} & 40.93\textcolor{gray}{\tiny{1.67}}\\

PlanTF    & 84.27 & 83.02\textcolor{gray}{\tiny{1.23}} & 68.83\textcolor{gray}{\tiny{0.96}} & 51.97\textcolor{gray}{\tiny{2.93}} & 79.29\textcolor{gray}{\tiny{1.27}} & 54.46\textcolor{gray}{\tiny{2.33}} & 42.69\textcolor{gray}{\tiny{0.99}} & 34.00 & 31.92\textcolor{gray}{\tiny{1.19}} & 29.76\textcolor{gray}{\tiny{4.53}} & 29.23\textcolor{gray}{\tiny{0.97}} & 31.93\textcolor{gray}{\tiny{0.35}} & 27.61\textcolor{gray}{\tiny{1.88}} & 23.07\textcolor{gray}{\tiny{1.70}}  \\

PLUTO     & 92.88 & 90.39\textcolor{gray}{\tiny{2.08}} & 79.77\textcolor{gray}{\tiny{3.59}} & 67.87\textcolor{gray}{\tiny{2.19}} & 87.76\textcolor{gray}{\tiny{0.66}} & 72.08\textcolor{gray}{\tiny{2.74}} & 63.39\textcolor{gray}{\tiny{2.65}} & 36.74 & 35.47\textcolor{gray}{\tiny{1.17}} & 32.04\textcolor{gray}{\tiny{2.06}} & 30.66\textcolor{gray}{\tiny{3.01}} & 32.01\textcolor{gray}{\tiny{3.80}} & 31.53\textcolor{gray}{\tiny{1.80}} & 29.77\textcolor{gray}{\tiny{2.03}} \\

Diffusion planner & 89.87 & 87.22\textcolor{gray}{\tiny{0.61}} & 77.52\textcolor{gray}{\tiny{5.28}} & 63.40\textcolor{gray}{\tiny{7.53}} & 84.31\textcolor{gray}{\tiny{2.48}} & 68.03\textcolor{gray}{\tiny{7.14}} & 52.90\textcolor{gray}{\tiny{7.52}} & 37.49 & 36.82\textcolor{gray}{\tiny{0.35}} & 35.67\textcolor{gray}{\tiny{1.98}} & 32.97\textcolor{gray}{\tiny{1.75}} & 35.91\textcolor{gray}{\tiny{0.84}} & 33.26\textcolor{gray}{\tiny{1.79}} & 30.03\textcolor{gray}{\tiny{2.82}} \\

CaRL      & 93.87 & 93.31\textcolor{gray}{\tiny{0.30}} & 92.23\textcolor{gray}{\tiny{0.86}} & 88.40\textcolor{gray}{\tiny{3.04}} & 93.19\textcolor{gray}{\tiny{0.30}} & 92.39\textcolor{gray}{\tiny{0.77}} & 87.08\textcolor{gray}{\tiny{4.89}} & 72.57 & 70.69\textcolor{gray}{\tiny{0.27}} & 70.09\textcolor{gray}{\tiny{1.18}} & 69.57\textcolor{gray}{\tiny{0.93}} & 70.44\textcolor{gray}{\tiny{0.33}} & 69.89\textcolor{gray}{\tiny{0.50}} & 68.63\textcolor{gray}{\tiny{1.00}} \\

\midrule
\multicolumn{15}{c}{\textbf{Reactive}} \\
\midrule

PDM-Closed & 92.12 & 90.67\textcolor{gray}{\tiny{0.97}} & 88.01\textcolor{gray}{\tiny{1.03}} & 76.46\textcolor{gray}{\tiny{2.39}} & 89.96\textcolor{gray}{\tiny{1.57}} & 81.93\textcolor{gray}{\tiny{1.97}} & 70.24\textcolor{gray}{\tiny{4.89}} & 49.95 & 48.92\textcolor{gray}{\tiny{0.98}} & 48.72\textcolor{gray}{\tiny{0.83}} & 46.79\textcolor{gray}{\tiny{1.60}} & 49.05\textcolor{gray}{\tiny{1.14}} & 48.20\textcolor{gray}{\tiny{3.35}} & 44.96\textcolor{gray}{\tiny{1.38}} \\

PlanTF    & 76.95 & 73.97\textcolor{gray}{\tiny{1.77}} & 64.96\textcolor{gray}{\tiny{1.77}} & 55.12\textcolor{gray}{\tiny{1.56}} & 70.06\textcolor{gray}{\tiny{0.45}} & 55.90\textcolor{gray}{\tiny{2.76}}  & 38.70\textcolor{gray}{\tiny{7.06}} & 49.44 & 48.39\textcolor{gray}{\tiny{0.64}} & 46.91\textcolor{gray}{\tiny{2.09}}  & 45.61\textcolor{gray}{\tiny{2.14}} & 46.38\textcolor{gray}{\tiny{1.60}} & 44.05\textcolor{gray}{\tiny{3.13}} & 37.51\textcolor{gray}{\tiny{3.28}} \\

PLUTO     & 92.06 & 85.96\textcolor{gray}{\tiny{2.95}} & 82.73\textcolor{gray}{\tiny{1.01}} & 72.60\textcolor{gray}{\tiny{2.84}} & 84.03\textcolor{gray}{\tiny{1.33}} & 74.61\textcolor{gray}{\tiny{4.68}} & 63.97\textcolor{gray}{\tiny{3.58}} & 63.03 & 62.38\textcolor{gray}{\tiny{1.91}} & 62.03\textcolor{gray}{\tiny{2.11}} & 60.93\textcolor{gray}{\tiny{0.60}} & 60.26\textcolor{gray}{\tiny{2.66}} & 59.91\textcolor{gray}{\tiny{3.01}} & 55.10\textcolor{gray}{\tiny{1.98}} \\

Diffusion planner & 82.80 & 81.22\textcolor{gray}{\tiny{0.50}} & 75.44\textcolor{gray}{\tiny{2.03}} & 65.59\textcolor{gray}{\tiny{3.94}} & 79.69\textcolor{gray}{\tiny{1.04}} & 68.78\textcolor{gray}{\tiny{3.09}} & 56.52\textcolor{gray}{\tiny{4.89}}& 49.39 & 49.34\textcolor{gray}{\tiny{0.29}} & 48.04\textcolor{gray}{\tiny{1.13}} & 46.61\textcolor{gray}{\tiny{1.66}} & 48.66\textcolor{gray}{\tiny{0.61}} & 47.38\textcolor{gray}{\tiny{2.24}} & 43.15\textcolor{gray}{\tiny{3.23}} \\

CaRL      & 93.12 & 92.97\textcolor{gray}{\tiny{0.33}} & 92.07\textcolor{gray}{\tiny{1.03}} & 88.30\textcolor{gray}{\tiny{2.54}} & 92.54\textcolor{gray}{\tiny{0.38}} & 91.14\textcolor{gray}{\tiny{1.56}} & 85.76\textcolor{gray}{\tiny{5.64}} & 70.27 & 69.91\textcolor{gray}{\tiny{0.55}} & 69.02\textcolor{gray}{\tiny{0.60}} & 67.88\textcolor{gray}{\tiny{0.49}} & 69.59\textcolor{gray}{\tiny{0.43}} & 68.18\textcolor{gray}{\tiny{0.56}} & 67.08\textcolor{gray}{\tiny{0.80}} \\
\bottomrule
\end{tabular}
}
\end{table*}

\subsection{Track 1: Semantic Shift}
\label{subsec:semantic_results}
Standard closed-loop benchmarks, such as nuPlan Val14, evaluate planners on unseen scenarios drawn from the same geographical regions as the training data. 
In this within-domain regime, architecturally diverse methods, from rule-based controllers to diffusion transformers, often converge to a narrow performance band (Table~\ref{tab:semantic_shift}, Val14). 
Our zero-shot benchmark exposes fundamental differences in generalization that these standard evaluations conceal.
Table~\ref{tab:semantic_shift} reports the CLSs for our novel scenarios in Sindelfingen (SIFI), Munich (MUC), Stuttgart (STR), Berlin (BER), San Francisco (SFO) and a metric breakdown of the overall CLS. 

\noindent\textbf{Results and Analysis.}
CaRL achieves the highest zero-shot performance with a 72.57 CLS, outperforming the next best baseline (PDM-Closed) by 24.5 points.

A possible explanation of its robustness can be identified into three fundamental differences in training methodology: 
(i) compared to IL that seeks to minimize a divergence from expert trajectories, CaRL optimizes for a reward signal (progress, safety, and comfort) that is directly aligned with the nuPlan evaluation metrics, making it inherently region-agnostic.
(ii) Furthermore, the stochastic exploration required during RL training forces the agent to visit and recover from off-policy states, effectively acting as a form of implicit data augmentation. 
(iii) CaRL is trained in both NR and R simulations exposing the policy to multiple behaviors.

IL methods exhibit severe sensitivity to semantic shift.
Diffusion Planner’s CLS collapses to 21.50 in Munich, where complex pedestrian interactions absent from the nuPlan training distribution dominates the scenario set.
This represents a 76\% degradation from its Val14 baseline (89.87).
PlanTF’s CLS drops from 84.27 (Val14) to 34.00, making it the worst model in terms of progress (PER).
PLUTO, which augments IL with a rule-based post-processing layer, achieves comparable overall CLS to Diffusion Planner in NR mode but exhibits an intriguing improvement under R simulation.
Its CLS increases from 36.74 (NR) to 63.03 (R), a 71.5\% improvement, whereas PDM-Closed and PlanTF improve by only $\sim$4-15 points.
This suggests that PLUTO's rule-based refinement module benefits from IDM-controlled agents that may provide larger safety margin compared to human log replay, effectively compensating for its IL backbone's distributional mismatch.

PDM-Closed, the purely rule-based planner, occupies a revealing middle ground. 
While it shares RL’s goal-oriented objective, it lacks the functional flexibility to navigate dense, non-linear social interactions. 
It avoids the catastrophic collisions of IL, reaching the highest safety score (NCR) among the evaluated planners (89.25 in NR and 90.10 in R).
However, it suffers from low progress with a PER of ~50 points, reflecting a tendency to prioritize safety margins. 

This demonstrates that our benchmark probes planning capabilities beyond what rule-based heuristics can provide, such as adaptive social reasoning in dense interactions.
A broader takeaway is that the failure modes differ fundamentally across paradigms: IL planners more frequently exhibited safety-critical failures (low NCR, low DDC), rule-based planners fail conservatively (low progress, high safety), and the RL planner degrades gracefully, maintaining both safety and progress under significant distributional shift.
The difference of behaviors and some challenging DeepPlan scenarios are exemplified in Fig.~\ref{fig:track1_qualitative}.

\subsection{Track 2: State-Distribution Drift}
\label{subsec:robustness_results}

While Track 1 evaluates generalization across environments, Track 2 tests resilience against execution-time perturbations. 
On the noise-free Val14 set, most planners appear equally capable. 
However, the introduction of actuation noise reveals a significant divergence in control stability (Table~\ref{tab:noise_comparison}).

\begin{figure}[tb]
    \centering
    \includegraphics[width=0.95\linewidth]{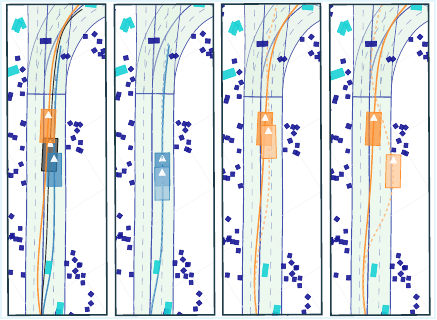}
    \caption{
    \textbf{Qualitative analysis of trajectory stability under actuation perturbations.} 
    From left to right: (1) Reference trajectories for \textbf{\textcolor{mylog}{Log}}, \textbf{\textcolor{mycarl}{CaRL}}, and \textbf{\textcolor{mydiff}{Diffusion Planner}} under nominal conditions; (2) CaRL under nominal conditions (solid) vs. high-intensity OU drift (dashed); (3) Diffusion Planner under nominal conditions (solid) vs. high-intensity AWGN jitter (dashed); (4) Diffusion Planner under nominal conditions (solid) vs. high-intensity OU drift (dashed).
    }
    \vspace{-1em}
    \label{fig:qual_noise}
\end{figure}

\noindent\textbf{Results and Analysis.} 
CaRL maintains high resilience: under AWGN and OU noise, it retains all CLSs above 85, which is a minimal drop compared to the other planners.

This robustness likely stems from the stochastic nature of RL training: by optimizing under exploration noise, the policy internalizes the vehicle's dynamic constraints and learns to recover from off-policy states.
Consequently, CaRL functions as a dynamics-aware governor that can adjust its control strategy to counteract drift.

Despite being also trained to recover from perturbed states through data augmentation, IL methods degrade sharply, and the pattern is noise-type-dependent. 
Under AWGN, which adds independent perturbations at each timestep, planners can partially self-correct on subsequent steps.
However, the CLS performance still significantly decrease by up to 20-30 points on the high noise settings.
Under the temporally correlated OU process errors accumulate as a persistent drift that IL policies, trained to match expert state-action pairs without closed-loop correction, cannot recover from. 
PlanTF and Diffusion Planner CLS falls by approximately an additional 10 points compared to Gaussian noise, highlighting a failure to recover from multi-frame errors.

Fig.~\ref{fig:qual_noise} shows a characteristic failure: the Diffusion Planner, subjected to OU noise, gradually drifts from the lane center over several seconds before crossing into oncoming traffic, while CaRL remains centered throughout.

PDM-Closed maintains its position as a middle-ground performer.
Its rule-based lane-following logic allows it to better compensate for deviations, preventing the drift-to-collision failures of IL methods. 
While safer than pure IL, this intermediate position highlights a fundamental limitation of reactive rule-based control under noisy actuation.

The most revealing result emerges from combining the two benchmark tracks. 
On the DeepPlan set (right half of Table~\ref{tab:noise_comparison}), noise compounds the generalization challenge non-additively: planners already struggling with the semantic shift see their remaining performance eroded further by actuation noise. 
CaRL's advantage is even more pronounced in this combined-stress regime, suggesting that RL-trained robustness generalizes across both distributional and dynamic perturbations simultaneously.

\section{Conclusions}
\label{sec:conclusions}

In this work, we introduced \textit{Shift \& Drift}, a novel two-track benchmark evaluating AD planners beyond conventional i.i.d. assumptions.
By integrating high-fidelity aerial datasets into a standard simulation framework and introducing actuation noise, we expose the significant fragility of SOTA motion planners under semantic shifts and state-distribution drifts. 

Our empirical analysis yields three critical insights:
\begin{itemize}
\item \textbf{The Generalization Gap:} We quantify the generalization capabilities of IL-based models.
Despite high fidelity in training domains, they exhibit brittleness in novel environments, often failing to negotiate dense pedestrian-cyclist interactions.
\item \textbf{The Robustness Paradox:} Pure IL paradigms lack reactive closed-loop correction mechanisms. 
Thus, they fail to recover from temporally correlated execution errors (e.g., OU noise), exposing control instabilities entirely masked by standard noise-free evaluations.
\item \textbf{Functional Flexibility vs. Rule-Based Heuristics:} Purely rule-based planners fail despite structurally aligning with the evaluation metrics, indicating manual heuristics are overly rigid for complex interactions. 
The evaluated RL agent maintained higher robustness through learned adaptability in complex environments where manual heuristics become excessively restrictive.
Disentangling the contributions of reward design, closed-loop training, and policy flexibility remains an open challenge for future work.
\end{itemize}

Ultimately, \textit{Shift \& Drift} provides the community with a rigorous framework for stress-testing social intelligence and dynamic stability of object-level planners. 
Future work will focus on scaling the semantic shift track to include a wider variety of global driving cultures, isolating the individual contributions of topology, interaction density, and regional driving norms to the observed performance degradation, and investigating hybrid architectures that combine the safety guarantees of rule-based systems with the resilient exploration of RL.


\bibliographystyle{IEEEtran}
\bibliography{IEEEabrv, related_work}

\end{document}